\documentclass[twocolumn,a4paper,fleqn]{cas-dc}

\usepackage[numbers]{natbib}
\usepackage{color}

\usepackage{amsmath,amssymb,amsfonts}
\usepackage{graphicx}
\usepackage{textcomp}
\usepackage{algorithmicx}
\usepackage[ruled,vlined]{algorithm2e}
\usepackage{algpseudocode}
\usepackage{amsmath}
\usepackage{amsmath,amssymb,amsfonts}
\usepackage[mathscr]{eucal}
\usepackage{graphicx}
\usepackage{textcomp}
\usepackage{graphicx}
\usepackage{float}
\usepackage{subfig}
\usepackage{url}
\usepackage{longtable}
\usepackage{booktabs}
\usepackage{multirow}

\usepackage{graphicx}
\usepackage{hyperref}
\hypersetup{hidelinks,
	colorlinks=true,
	allcolors=blue,
	pdfstartview=Fit,
	breaklinks=true}

\begin{document}
\let\WriteBookmarks\relax
\def\floatpagepagefraction{1}
\def\textpagefraction{.001}
\shorttitle{Bundle-specific Tractogram Distribution Estimation}

\title [mode = title]{Bundle-specific tractogram distribution estimation using higher-order streamline differential equation}

\author%
[1,2,3]
{Yuanjing~Feng}
\cormark[1]
\cortext[cor1]{Corresponding author}
\ead{fyjing@zjut.edu.cn}
\address[1]{College of Information Engineering, Zhejiang University of Technology, Hangzhou, China}
\author[1,2,3]{Lei~Xie}
\author[1]{Jingqiang~Wang}
\author%
[4]
{Qiyuan~Tian}
\cormark[1]
\ead{qiyuantian@tsinghua.edu.cn}
\author[1,2,3]{Jianzhong~He}
\author[1,2,3]{Qingrun~Zeng}
\author[1]{Fei~Gao}
\address[2]{Zhejiang Provincial Collaborative Innovation Center for High-end Digital Intelligence Diagnosis and Treatment Equipment, Hangzhou, China}
\address[3]{Zhejiang Provincial United Key Laboratory of Embedded Systems, Hangzhou, China}
\address[4]{Department of Biomedical Engineering, Tsinghua University, Beijing, China}

\begin{abstract}
Streamline tractography locally traces peak directions extracted from fiber orientation distribution (FOD) functions, lacking global information about the trend of the whole fiber bundle. Therefore, it is prone to producing erroneous tracks while missing true positive connections. In this work, we propose a new bundle-specific tractography (BST) method based on a bundle-specific tractogram distribution (BTD) function, which directly reconstructs the fiber trajectory from the start region to the termination region by incorporating the global information in the fiber bundle mask. A unified framework for any higher-order streamline differential equation is presented to describe the fiber bundles with disjoint streamlines defined based on the diffusion vectorial field. At the global level, the tractography process is simplified as the estimation of BTD coefficients by minimizing the energy optimization model, and is used to characterize the relations between BTD and diffusion tensor vector under the prior guidance by introducing the tractogram bundle information to provide anatomic priors. Experiments are performed on simulated Hough, Sine, Circle data, ISMRM 2015 Tractography Challenge data, FiberCup data, and in vivo data from the Human Connectome Project (HCP) for qualitative and quantitative evaluation. Results demonstrate that our approach reconstructs complex fiber geometry more accurately. BTD reduces the error deviation and accumulation at the local level and shows better results in reconstructing long-range, twisting, and large fanning tracts.
\end{abstract}




\maketitle
\section{Introduction}
\label{sec:introduction}
\begin{figure}[t]
	\centering
	\includegraphics[width=0.49\textwidth]{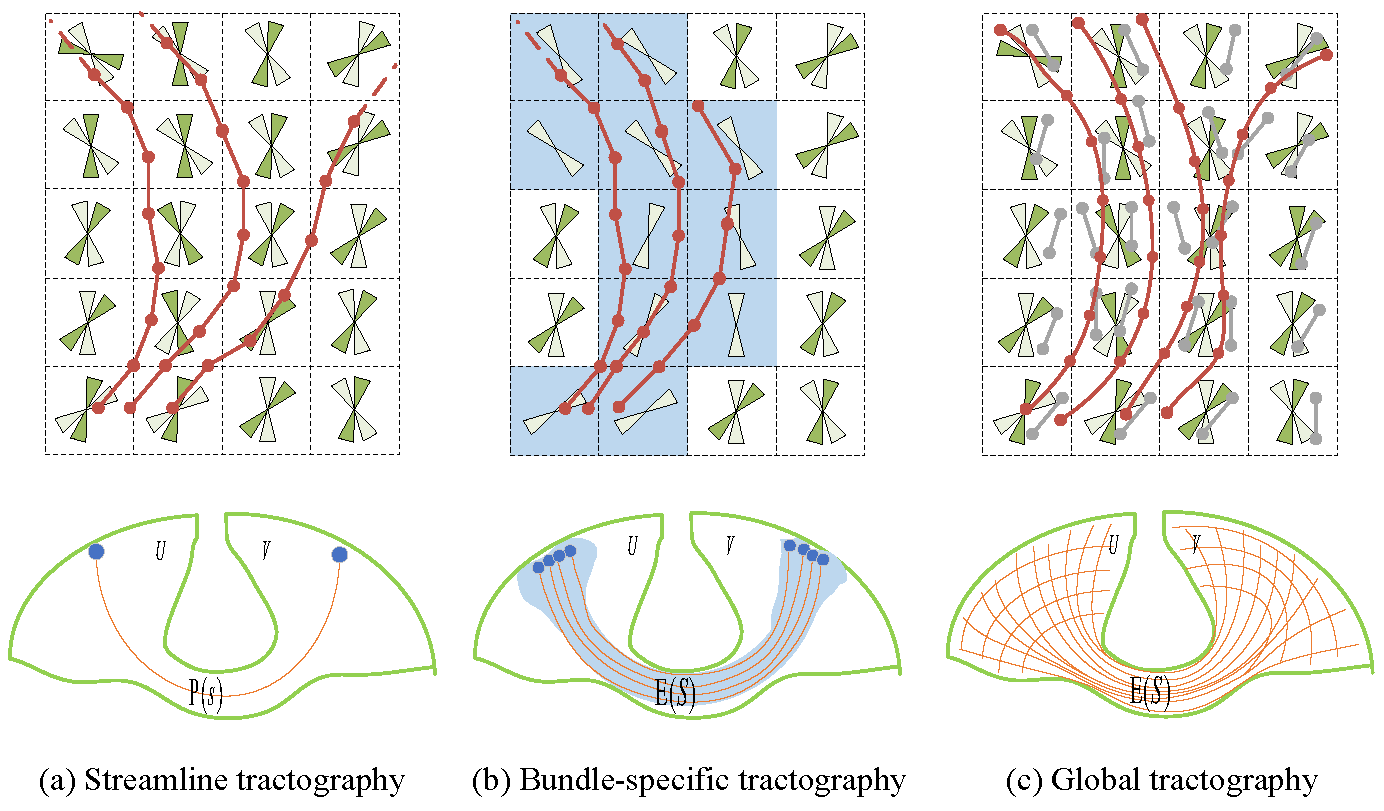}
	\caption{Schematic representation of different tractography methods.}
	\label{fig:1}
\end{figure}

Diffusion magnetic resonance imaging (dMRI) tractography (i.e., fiber tracking) is a useful tool for delineating white matter (WM) pathways and reconstructing structural connectome non-invasively in the in vivo human brain~\cite{1,2,3,32,li2014knowledge}. Typical tractography algorithms propagate streamlines through the vector field of fiber orientations extracted from fiber orientation distribution (FOD) functions estimated at each image voxel to reconstruct long-range connections~\cite{4}. This type of method is therefore often referred to as streamline tractography (ST, Fig.~\ref{fig:1}a) and is widely known to produce false-positive fibers and/or omit true-positive fibers~\cite{5,6,7}, due to the ambiguous spatial correspondences between diffusion orientations and fiber geometry on the voxel-wise, local and global level~\cite{1,8,9}. Deterministic ST algorithms initiate each streamline from a seed voxel and then follow the peaks extracted from FODs to the termination region, which inevitably accumulates errors in the propagation process. Rather than using fixed fiber orientation(s), probabilistic ST algorithms randomly draw a direction based on the reconstructed FOD at each image voxel and perform the tracking many times using an iterative Monte Carlo approach, mapping all the possible pathways while producing a large number of false-positive fibers in the meanwhile~\cite{19,20,21,60}. To reduce the generation of false-positive fibers, ST algorithms typically combine regions of interest (ROIs) selection strategies~\cite{wakana2007reproducibility} or voxel-wise atlases~\cite{zhang2010atlas} to analyze the generated fiber streamlines conform to anatomical regions.

Numerous signal modeling and tracking techniques have been proposed to improve the performance of streamline tractography. Early efforts focus on resolving multiple fibers within a voxel to solve the major problem of diffusion tensor imaging (DTI) which cannot resolve crossing and kissing fibers~\cite{tuch2004q,10,11}. Nonetheless, these models still cannot accurately map fanning and bending fibers with asymmetric curvature in the range of the voxel size~\cite{3,12}. Asymmetric FODs (AFODs) are consequently proposed to address this challenge~\cite{12,13,14}. A set of approaches leveraged local surrounding voxels and intervoxel correlation to characterize underlying fiber patterns ~\cite{15,16,17}. Unfortunately, AFODs only reduce the spatial ambiguity on the voxel and local level. Simply plugging them into a streamline tractography algorithm is not useful in mitigating the ambiguity on the global level for mapping complex tractograms. Additionally, many sophisticated tracking strategies have also been leveraged. For instance, “filtered tractography” reduces tracking biases by using an unscented Kalman filter to simultaneously fit the local the orientation distribution functions (ODF) model to diffusion signals and propagate a streamline in the most consistent direction ~\cite{22,23}. “Parallel transport tractography” \cite{30} generates geometrically smooth curves using the more flexible parallel transport frames and topographic regularity of connections. Nevertheless, the optimization only happens on the local level, without knowing any global information of the fiber trajectory. 

Global tractography (GT)~\cite{26} methods aim to simultaneously reconstruct all tracks that can best describe the measured dMRI data (Fig.~\ref{fig:1}c). In this way, the estimation of local fiber orientations at each voxel is informed by surrounding voxels ~\cite{25,27,28}. For example, Jbabdi et al.~\cite{24} proposed a GT framework to guide probabilistic tracking, which can infer local directional parameters from large-scale connectivity in the presence of high local data uncertainty, based on a priori knowledge about regional correlations. GT offers improved agreement with the measured dMRI data as well as increased robustness to noise and imaging artifacts~\cite{jeurissen2019diffusion}. Even though GT algorithms map crossing fiber regions slightly more accurately, resultant bundle flow maps do not always represent the known fascicular topology~\cite{li2012quantitative}. The complexity of GT can be improved, at a cost of significantly increased computational burden, which is unacceptable or impossible in many applications. 

Another category of tracking method is the recently proposed “bundle-specific tractography” (BST)~\cite{rheault2018bundle}. Rather than mapping each streamline independently in ST or all streamlines simultaneously in GT, BST works on each fiber bundle of interest. Consequently, BST can incorporate anatomical and orientational prior knowledge of the tract into the tracking process (Fig.~\ref{fig:1}b). Specifically, BST first assigns fiber populations (e.g., peaks from the FOD) within a voxel to different fiber tracts, using a neural network ~\cite{wasserthal2018tract,33} or an atlas, and performs classic streamline tractography only using the assigned fiber populations for each tract separately. BST results are more anatomically plausible, with more accurate delineation of intersected fiber tracts and larger spatial coverage to represent the full shape of bundles~\cite{58}. Nonetheless, current BST methods still use conventional deterministic or probabilistic tracking and suffer from pitfalls of streamline tractography.

To address this challenge, we propose to reconstruct all streamlines between the seeding and targeting region for a fiber tract of interest simultaneously, which is essentially semi-global or tract-wise global tractography. Specifically, we define a new bundle-specific tractogram distribution (BTD) function to describe the tract-wise tractogram based on any higher-order streamline differential equation in the measured diffusion tensor vector field. The optimization model is reconstructed with the measured diffusion orientations and BTD function. At the global level, the tractography process is to parameterize as BTD coefficients combining voxel location by minimizing the energy of the optimization model. The relations between BTD and diffusion tensor vector are described under the prior guidance by introducing the tractography atlas as anatomic priors. 

Our proposed method is validated on three simulated datasets (Hough dataset, Sine dataset, and Circle dataset), the ISMRM 2015 Tractography Challenge dataset,  FiberCup dataset, and the Human Connectome Project (HCP) ~\cite{34,35} in vivo dataset. Results show that BTD can reconstruct complex fiber bundles, such as long-range, large twisting, and fanning tracts, and show better spatial consistency with fiber geometry. An open-source implementation of BTD is available at~\href{https://github.com/IPIS-XieLei/BTD-Tractography}{https://github.com/IPIS-XieLei/BTD-Tractography}. Our contributions are summarized as follows:
\begin{enumerate}[(1)]
	\item [i)] A novel BTD function for fiber tractography to directly reconstruct the fiber trajectory is proposed.
	\item [ii)] A unified framework for any higher-order streamline differential equation is presented to describe the fiber bundles with disjoint streamlines defined based on the diffusion tensor vector field.
	\item [iii)] The fiber bundles are parameterized using the BTD coefficients, which are estimated by combining the priors and minimizing the energy on the diffusion tensor vector field.
	\item [iv)] Experimental results on Hough, Sine, Circle, FiberCup, ISMRM 2015 data, and HCP dataset show that the BTD is capable of reconstructing complex fiber bundles with long distances, large twists, and fan-shaped bundles, and shows better spatial consistency with the fiber geometry.
\end{enumerate}

The rest of the paper is organized as follows. Section~\ref{sec:Methods} defines the BTD based on a higher-order streamline differential equation. In Section~\ref{sec:Experiments}, the experimental results on six datasets are presented. Section~\ref{sec:Discussion} discusses the results and future works, and Section ~\ref{sec:Conclusion} gives a summary of the proposed work.

\section{Methods}
\label{sec:Methods}
In its most basic form, a tractography algorithm takes two arbitrary points (voxels) of interest as input, labeled as a seed and a target, and yields the most likely trajectory on the given diffusion vectorial field. Let $\mathcal{D}  \subset {R^3}$ be the diffusion vectorial field, and ${S_{U,V}}(t) \subset \mathcal{D}$ denotes a path that is parameterized by $t \in [0,T]$ connecting $U$ with $V$. Let $\mathrm{P}(s)$ denote the probability of paths $s$ (Fig.~\ref{fig:1}a) representing an anatomically genuine fiber trajectory in diffusion native space $\mathcal{D}  \subset {R^3}$, which can be defined as,
\begin{equation}
	\mathrm{P}(s) = \int_0^T {p(s(t),\dot s(t))} {\rm{ }}dt,
	\label{eq1}
\end{equation}
where $s(0) = {t_1}$, $s(T) = {t_2}$ and $p\left( {s(t),\dot s(t)} \right)$ is the metric representing the potential for the steamline $s(t)$ from two arbitrary points ${t_1}$ and ${t_2}$ to be located inside a fiber bundle in the direction $\dot s(t) = {{ds(t)} \mathord{\left/{\vphantom {{ds(t)} {dt}}} \right.\kern-\nulldelimiterspace} {dt}}$. In an early deterministic tracking algorithm, (streamline tractography), $s(t)$ is computed to satisfy the Frenet equation,
\begin{equation}
	\frac{{{\rm{d}}s(t)}}{{dt}} = {V_{\max }}(s(t)),t \in \left[ {0,T} \right],
	\label{eq2}
\end{equation}
where ${V_{\max }}(s(t))$ is the principal diffusion direction of $s(t)$. In a sense, this is a "greedy" algorithm, as it tries to find the optimal fiber trajectory. While shortest path methods aim to find an optimal path by computing the maximum energy between two seeds. 

In this paper, we are interested in the particular case of computing the diffusion vectorial field on a Riemannian manifold~\cite{41}, which is a potential under the form $p\left( {s(.),\dot s(.)} \right) = \sqrt {\dot s{{(.)}^T}{\rm M}(s(.))\dot s(.)}$ describing an infinitesimal distance along the fiber path $S$ relative to the metric tensor $M$ (symmetric definite positive), which is usaually defined by the diffusion tensor~\cite{basser1994mr}. In this situation, finding the curve connecting two points that globally minimizes the energy~\cite{42} is a shortened path called a geodetic~\cite{43,44}. In~\cite{45}, a variational model is proposed based on Hamilton-Jacobi-Bellman, in which an infinite number of particles start from a given seed region evolving along the streamline orientation given by the gradient of the defined cost function to reach the seed target. In our recent work~\cite{16,31}, penalized geodesic tractography based on a Finsler metric with a global optimization framework is introduced to improve cortical connectomics. These methods are related to an optimal control problem and turn out to be very useful for establishing the connectivity of a single point on the seed region but provide little information about the connectivity between two regions. Daducci et al. \cite{46} attempted to introduce optimal mass transportation to describe the connectivity between two cerebral regions. Unfortunately, it is only a preliminary mathematical description without algorithmic implementation and validation. 

Herein, we will focus here on the bundle structure tractography connecting two cerebral regions in a "cluster to cluster" manner. Considering two given regions ${\rho _1}$ and ${\rho _2}$, the optimal fiber bundle can be viewed as a superposition of non-intertwined fiber streamline cluster $S(t)$ starting from ${\rho _1}$ to ${\rho _2}$ along the diffusion vectorial field, and the energy $\mathrm{E}(S)$ assigned to the streamline cluster is defined as, 
\begin{equation}
	\mathrm{E}(S) = \int_{\rm{0}}^T {p\left( {S(t),\dot S(t)} \right)} {\rm{ }}dt,
	\label{eq3}
\end{equation}
where $S(0) = {\rho _1}$, $S(T) = {\rho _2}$ and $\dot S(t)$ is the diffusion vectorial field between ${\rho _1}$ and ${\rho _2}$. We first define the $S(t)$ and  $\dot S(t)$, and $p\left( {S(t),\dot S(t)} \right)$. In Section~\ref{sec:a}, we define the BTD function $S(t)$ on the diffusion vectorial field $\dot S(t)$ from ${\rho _1}$ to ${\rho _2}$ based on a higher-order streamline differential equation. For bundle continuity, we add spatial continuity constraint equation in Section~\ref{sec:b}. Then, the $S(t)$ solution can be simplified as estimation of BTD coefficients, detailed process is shown in Section~\ref{sec:c}.
\subsection{Bundle-specific tractogram distribution function}\label{sec:a}
Let the diffusion vectorial field at position $(x,y,z)$ be
\begin{equation}
	v(x,y,z) = {\left[ {{v_x}(x,y,z),{v_y}(x,y,z),{v_z}(x,y,z)} \right]^T}.
	\label{eq4}
\end{equation}
The BTD can be parameterized by a 3D fiber bundle with a set of streamlines $S(t) = \left\{ {{s_i},i = 1, \cdots ,\lambda } \right\}$ that satisfies the following properties~\cite{31}:

i) The tangent vector at each point $(x,y,z)$ of fiber path ${s_i}$ equals the field vector $v(x,y,z)$, that is,
\begin{equation}
	\frac{{\partial {s_i}}}{{\partial x}}{\rm{  }} = {\rm{ }}{v_x}(x,{\rm{ }}y,{\rm{ }}z){\rm{ }},{\rm{ }}\frac{{\partial {s_i}}}{{\partial y}}{\rm{ }} = {\rm{ }}{v_y}(x,{\rm{ }}y,{\rm{ }}z){\rm{ }},{\rm{ }}\frac{{\partial {s_i}}}{{\partial z}} = {\rm{ }}{v_z}(x,{\rm{ }}y,{\rm{ }}z){\rm{ }}{\rm{.}}
	\label{eq5}
\end{equation}

ii) The streamlines satisfy ${s_i} \cap {s_j} = 0$, $i \ne j$, which is defined using the streamline differential equation,
\begin{equation}
	\frac{{dx}}{{{v_x}(x,{\rm{ }}y,{\rm{ }}z)}}{\rm{ }} = {\rm{ }}\frac{{dy}}{{{v_y}(x,{\rm{ }}y,{\rm{ }}z)}} = \frac{{dz}}{{{v_z}(x,{\rm{ }}y,{\rm{ }}z)}}.
	\label{eq6}
\end{equation}

We introduce the higher-order streamline differential Eq.~\ref{eq7} with the tangent vector approximated by the ${n^{th}}$-order polynomial, 
\begin{equation}
	f(x,y,z) = \sum\limits_{i = 0}^n {\sum\limits_{j = 0}^{n - i} {\sum\limits_{k = 0}^{n - i - j} {a_{ijk}^{}{x^i}{y^j}{z^k}} } },
	\label{eq7}
\end{equation}
where ${a_{ijk}}$ is the coefficients of the polynomial. Combined with Eq.~\ref{eq4} and Eq.~\ref{eq7}, the diffusion vector $v(x,y,z)$ is denoted,
\begin{equation}
	\begin{array}{l}
		v(x,y,z){\rm{ = }}{\left[ \begin{array}{l}
				\sum\limits_{i = 0}^n {\sum\limits_{j = 0}^{n - i} {\sum\limits_{k = 0}^{n - i - j} {a_{ijk}^x{x^i}{y^j}{z^k}} } } \\
				\sum\limits_{i = 0}^n {\sum\limits_{j = 0}^{n - i} {\sum\limits_{k = 0}^{n - i - j} {a_{ijk}^y{x^i}{y^j}{z^k}} } } \\
				\sum\limits_{i = 0}^n {\sum\limits_{j = 0}^{n - i} {\sum\limits_{k = 0}^{n - i - j} {a_{ijk}^z{x^i}{y^j}{z^k}} } } 
			\end{array} \right]^T}\\
		{\rm{             = }}A \cdot C(x,y,z){\rm{ }}
	\end{array}
	\label{eq8}
\end{equation}
where $A$ is the coefficient matrix defined as, 
\[A = {\left[ \begin{array}{l}
	a_{n00}^x{\rm{  }}a_{(n - 1)10}^x{\rm{  }}a_{(n - 1)01}^x{\rm{  }}a_{(n - 1)00}^x{\rm{   }}...{\rm{  }}a_{001}^x{\rm{ }}a_{000}^x\\
	a_{n00}^y{\rm{  }}a_{(n - 1)10}^y{\rm{  }}a_{(n - 1)01}^y{\rm{  }}a_{(n - 1)00}^y{\rm{   }}...{\rm{   }}a_{001}^y{\rm{ }}a_{000}^y\\
	a_{n00}^z{\rm{  }}a_{(n - 1)10}^z{\rm{  }}a_{(n - 1)01}^z{\rm{  }}a_{(n - 1)00}^z{\rm{   }}...{\rm{   }}a_{001}^z{\rm{ }}a_{000}^z
	\end{array} \right]_{\left( {3\sum\limits_{\Delta  = 0}^n {\frac{{(\Delta  + 1)*(\Delta  + 2)}}{2}} } \right)}}\]
and $C(x,y,z)$ denotes as,
	\begin{equation}
		\begin{array}{l}
			{C_n}(x,y,z) = {({[{x^n},{x^{n - 1}}y,{x^{n - 1}}z,{x^{n - 1}}{\rm{, }} \cdot  \cdot  \cdot ,z{\rm{,}}1{\rm{ }}]^T})_{\left( {\sum\limits_{\Delta  = 0}^n {\frac{{(\Delta  + 1)*(\Delta  + 2)}}{2}} {\rm{,1}}} \right)}}\\
			{\rm{                 = [}}{{\rm{F}}_n}(x,y,z){\rm{, }}{C_{n - 1}}(x,y,z)],
		\end{array}
		\label{eq9}
	\end{equation}
	where ${C_1}(x,y,z) = [x,y,z,1]{\rm{ }}$; ${F_n}(x,y,z) = [{c_m},\cdot\cdot\cdot{\rm{ }},{c_l},\cdot\cdot\cdot{\rm{, }}{c_1}]$ with $m = (n + 1)(n + 2)/2$ and ${c_l} = {x^i}{y^j}{z^k}$ where $l,i,j,k$ satisfy with $l = i(n - \frac{{i - 3}}{2}) + j + 1$, $i = [0,1, \cdot  \cdot  \cdot ,n]$, $j = [0,1, \cdot  \cdot  \cdot ,n - i]$, and $i + j + k = n$.

To further illustrate the Eq.~\ref{eq9}, we give an example of ${C_n}$ with $n=2$, The $n$ represented the order of higher-order streamline differential equation
\[\begin{array}{l}
{C_2}(x,y,z){\rm{ = [}}{{\rm{F}}_2}(x,y,z){\rm{, }}{C_1}(x,y,z)]\\
{\rm{             }} = [{x^2},{y^2},{z^2},x{y^{}},x{z^{}},yz,x,y,z,{1^{}}]\\
\left\{ \begin{array}{l}
{C_1}(x,y,z) = [x,y,z,1]{\rm{ }}\\
{F_2}(x,y,z) = [{c_6},{c_5},{c_4},{c_3},{\rm{ }}{c_2}{\rm{, }}{c_1}]{\rm{ with}}\\
{\rm{ }}{c_6} = {x^2},{c_5} = {y^2},{c_4} = {z^2},{c_3} = x{y^{}},{c_{12}} = x{z^{}},{c_1} = y{z^{}}
\end{array} \right.
\end{array}\]

\subsection{Spatial continuity constraint equation for a tractogram}\label{sec:b}
We assume that the diffusion displacement of water molecules in the same fiber maintains continuity. We use continuous incompressible fluid theory to describe the spatial continuity of the bundle by introducing the concept of divergence of the fiber flow on diffusion vectorial field. Let $div\Gamma$ denote the divergence of the fiber flow on diffusion vectorial field, which is formulated as
\begin{equation}
	div\Gamma = \frac{{\partial {v_x}}}{{\partial x}} + \frac{{\partial {v_y}}}{{\partial y}} + \frac{{\partial {v_z}}}{{\partial z}},
	\label{eq10}
\end{equation}
where $v_x$, $v_y$, and $v_z$ represent vector value in $x$, $y$, and $z$ axis. We assume that the fibers do not originate or terminate in the white matter, that is, $div\Gamma$ satisfies
\begin{equation}
	div\Gamma = 0.
	\label{eq11}
\end{equation}
The substitution of Eq.~\ref{eq8}, and Eq.~\ref{eq10} into Eq.~\ref{eq11} yields:
\begin{equation}
	\begin{array}{l}
		div\Gamma = \sum\limits_{i = 0}^n {\sum\limits_{j = 0}^{n - i} {\sum\limits_{k = 0}^{n - i - j} {\left( \begin{array}{l}
						a_{ijk}^xi{x^{i - 1}}{y^j}{z^k} + \\
						a_{ijk}^yj{x^i}{y^{j - 1}}{z^k} + \\
						a_{ijk}^zk{x^i}{y^j}{z^{k - 1}}
					\end{array} \right)} } } {\rm{ }}\\
		{\rm{       = }}{H_{n - 1}} \cdot {C_{n - 1}}(x,y,z) = 0
	\end{array}
	\label{eq12}
\end{equation}
where ${C_{n - 1}}(x,y,z)$ can be derived from Eq.~\ref{eq9}, and 
\[{C_{n - 1}}(x,y,z) = {[{x^{n - 1}},{x^{n - 2}}y,{x^{n - 2}}z, \cdot  \cdot  \cdot ,{x^i}{y^j}{z^k}, \cdot  \cdot  \cdot ,z,1]_{\left( M,1 \right)}}\]
\[{H_{n - 1}} = {\left[ \begin{array}{l}
	na_{n00}^x{\rm{ +   }}a_{(n - 1){\rm{1}}0}^y{\rm{ + }}a_{(n - 1)01}^z{\rm{ }}\\
	(n - 1)a_{(n - 1)10}^x{\rm{  +  2}}a_{{\rm{(n - 2)2}}0}^y{\rm{  +  }}a_{(n - 2)11}^z\\
	(n - 1)a_{(n - 1)01}^x{\rm{  +  }}a_{(n - 2)11}^y{\rm{  +  2}}a_{(n - 2)02}^z\\
	(n - 1)a_{(n - 1)00}^x{\rm{  +  }}a_{(n - 2)10}^y{\rm{  +  }}a_{(n - 2)01}^z\\
	\cdot  \cdot  \cdot \\
	{\rm{(}}i{\rm{ + 1)}}a_{(i + 1)jk}^x + (j + 1)a_{i(j + 1)k}^y + \\(k + 1)a_{ij(k + 1)}^z\\
	\cdot  \cdot  \cdot \\
	a_{101}^x{\rm{ +  }}a_{011}^y{\rm{ + 2}}a_{002}^z\\
	a_{100}^x{\rm{ +  }}a_{010}^y{\rm{ +  }}a_{001}^z
	\end{array} \right]^T}_{\left( 1,M \right)}\]
where $M=\sum\limits_{\Delta  = 0}^n {\frac{{(\Delta )*(\Delta  + 1)}}{2}}$. 
\subsection{Estimation of the BTD}\label{sec:c}
With the definition of BTD, finding the optimal tractogram from ${\rho _1}$ to ${\rho _2}$ can be simplified as the estimation of coefficient matrix $A$. The coefficient matrix $A$ is estimated by minimizing the energy $\mathrm{E}(S)$ in Eq.~\ref{eq3} with $p\left( {S(t),\dot S(t)} \right)$  defined as: 
\begin{equation}
	p\left( {S(t),\dot S(t)} \right) = \left\| {\Phi (v(x,y,z)) - v(x,y,z)} \right\|_2^2
	\label{eq13}
\end{equation}
where $\Phi (v(x,y,z))$ is the probability that fiber trajectory is actually passes through the FOD at point $(x,y,z)$, which is derived from the ODF in each voxel in the volume. To achieve fiber spatial continuity, we add constrain Eq.~\ref{eq12} to Eq.~\ref{eq3}, which yields the optimization model:
\begin{equation}
	\begin{array}{l}
		\mathop {\min }\limits_S \mathrm{E}(S) = \iiint_{\Omega}{\left\| {\Phi (v(x,y,z)) - v(x,y,z)} \right\|_2^2dxdydz
		} \\ 
		s.t.{\rm{ }}{H_{n - 1}} \cdot {C_{n - 1}}(x,y,z) = 0
	\end{array}
	\label{eq14}
\end{equation}
where $\Omega  = [{\omega _1},{\omega _2}, \cdot  \cdot  \cdot ,{\omega _\gamma }]$ is the bundle pathway containing $\gamma$ voxels from ${\rho _1}$ to ${\rho _2}$. For simplicity, we approximate $\Phi (v(x,y,z))$ as the peak direction values in each voxel and set $[G = {[{g_1}{\rm{, }}{g_2} \cdot  \cdot  \cdot {g_\gamma }]_{\left( {3,{\rm{ }}\gamma } \right)}}$ as a set of peak direction values in $\omega$. The $(X,Y,Z)$ is the center of a voxel in $\omega$ and $C_n^*(X,Y,Z)$ is a set of $C_n(X,Y,Z)$. The estimation of coefficient matrix $A$ of BTD is simplified as 
\begin{equation}
\begin{array}{l}
\mathop {\min }\limits_A E = \left\| {G - A \cdot C_n^*(X,Y,Z)} \right\|_2^2\\
s.t.{\rm{ }}{H_{n - 1}} \cdot C_{n - 1}^*(X,Y,Z) = 0\\
	\begin{split}
		C_n^*(X,Y,Z) = &[C_n^{}({X_1},{Y_1},{Z_1}),C_n^{}({X_2},{Y_2},{Z_2}), \\ &\cdot  \cdot  \cdot ,C_n^{}({X_\gamma },{Y_\gamma },{Z_\gamma })]_{\left( {\sum\limits_{\Delta  = 0}^n {\frac{{(\Delta  + 1)*(\Delta  + 2)}}{2}} ,{\rm{ }}\gamma } \right)}
	\end{split}
\end{array}
\label{eq15}
\end{equation}
The estimation of the BTD is decomposed into two stages. First, the least squares method is used to solve Eq.~\ref{eq15} to obtain the coefficient matrix $A$. Second, we use the coefficient matrix $A$ to obtain the flow field vectors and use the fourth-order Runge-Kutta method~\cite{26} to solve the higher-order streamline differential equation Eq.~\ref{eq7}. See Algorithm.~\ref{Algorithm:1} for a detailed implementation. The process for generating the bundle masks in our experiments as follows: i) In vivo data:
First, constrained spherical deconvolution (CSD) is applied to obtain the three principal FOD directions per voxel (Peaks) which is the input to the trained U-Nets \cite{58}. Second, the optimal weights of the trained U-Nets are loaded to predict the given bundle mask and the given start/end region mask for each tract. ii) The simulated data: The bundle mask used were provided in the these public dataset.
\begin{algorithm}[]
	\caption{Estimation of the bundle-specific tractogram distribution function.}
	\label{Algorithm:1}
	\KwIn{A set of peak directions $G = {[{g_1}{\rm{, }}{g_2} \cdot  \cdot  \cdot {g_\gamma }]_{\left( {3,{\rm{ }}\gamma } \right)}}$ , the 
		center coordinate $(X,Y,Z)$ of each voxel in bundle 
		mask $\Omega  = [{\omega _1},{\omega _2}, \cdot  \cdot  \cdot ,{\omega _\gamma }]$, seed points $\left\{ {{\beta _i},i = 1, \cdots ,\lambda } \right\}$, tractography step size $\Delta$.
	}
	\KwOut{Tractogram $S$.} 
	Calculate $C_n^*(X,Y,Z)$ and $C_{n-1}^*(X,Y,Z)$ using $(X,Y,Z)$ in $\Omega$ with Eq.~\ref{eq9}.\\
	Calculate $H_{n-1}$ according to $C_{n-1}^*(X,Y,Z)$ by Eq.~\ref{eq12}.\\
	Estimate coefficients matrix $A$ with Eq.~\ref{eq15}.\\
	Calculate tractogram $S = \left\{ {{s_i},i = 1, \cdots ,\lambda } \right\}$ in $\Omega$.\\
	Take seed ${\beta _i}$ as starting points of the tractogram $S$. \\
	\For{$i = 1 \to \lambda$}{
		set $\eta (0)$ as starting point of $S_i$ \\
		$\eta (0) = {\beta _i}$, $t=0$\\
		\While{$\eta (t) \in \Omega$}{
			Calculate flow field direction at point $\eta (t)$\\
			$v(\eta (t)) = A \cdot {C_n}(\eta (t))$\\
			Calculate the next point of $S_i$ with Eq.~\ref{eq6}\\
			$\eta (t + 1) = RungeKutta({\rm{ }}\eta (t),v(\eta (t)),\Delta )$\\
			${s_i} = [\eta (0), \cdots ,{\rm{ }}\eta (t)]$\\
			$t = t+1$\\
			
		}
		$S = \left\{ {{s_i},i = 1, \cdots ,\lambda } \right\}$	
	}
\end{algorithm} 
\section{Experiments}\label{sec:Experiments}
Tests and validations of our algorithm are based on four simulated datasets, namely, Hough data, Sine data~\cite{26}, Circle data, FiberCup data, and ISMRM 2015 Tractography Challenge data~\cite{1}, and in vivo data of HCP~\cite{35}. Our BTD estimation is based on FODs estimated using constrained spherical deconvolution, which are included in the software package MRtrix~\cite{10}. The experimental results are evaluated by comparing BTD with deterministic FOD-based tracking (SD\_Stream)~\cite{56}, integration over fiber orientation distributions (iFOD2)~\cite{57}, and unscented Kalman filter (UKF) algorithm~\cite{47}. For quantitative evaluation, we use Tractometer metrics, that is, valid connections (VC), bundle overlap (OL), and bundle overreach (OR) for Sine and Hough data~\cite{1,48,49}. For Circle data, we design a metric referred to as $Deviation$, which can be expressed as:
\begin{small} 
\begin{equation} 
Deviation = \sum\limits_{j = 1}^N {\sum\limits_{i = 1}^{{\lambda _j}} {(\sqrt {{{({x_{ij}} - {U_c})}^2} + {{({y_{ij}} - {V_c})}^2}} } }  - {R_j})/\sum\limits_{j = 1}^N {({\lambda _j})}, 
\end{equation}
\end{small}
where $N$ is the number of streamlines belonging to a bundle, and the $j$-th streamline has ${\lambda _j}$ points. The $({x_{ij}},{y_{ij}})$ is the $i$-th point on $j$-th streamline, and ${R _j}$ is the distance from a seed to circle center $(U_c,V_c)$ on $j$-th streamline. In addition, we compute the bundle volumetric overlap to quantify the spatial coverage~\cite{1}.
\subsection{Hough, Sine, Circle, and FiberCup data}
We simulate three datasets: the Hough data feature brain-like large fanning connections, while the Sine data feature long-range and large twisting connections, and the Circle data feature large bending connections. The datasets are generated with the following parameter settings. The spatial resolution is $1mm \times 1mm \times 1mm$. There are 78 gradient directions with a b-value of 1000 $s/m{m^2}$, and the signal-to-noise ratio (SNR) is set to 10, 20, and $+ \infty$. The Hough and Circle data size is $60 \times 60 \times 6$, and the Sine data size is $100 \times 100 \times 6$. Sine data satisfy ${\rm{2}}\pi  \cdot {\rm{y}} = \alpha  \cdot \sin (x)$, and $\alpha$ is set to 0.1, 0.2, 0.3, and 0.4 to adjust different amplitudes. The inner radius ($r_1$) of the circle data is 10 $mm$, and the outer radius ($r_2$) is 20 $mm$. FiberCup data were released in the MICCAI Challenge in 2009. The spatial resolution is $3mm \times 3mm \times 3mm$; the data have 30 gradient directions, with a b-value of 1000 $s/m{m^2}$ and a size of $64 \times 64 \times 3$.

\begin{table}[h]
	\centering
	\caption{Comparison of Tractometer metrics of BTD using Hough data and Sine data with SD\_Stream, and IFOD2.}
	\label{tab:1}
	\resizebox{0.46\textwidth}{!}{%
		\begin{tabular}{@{}cccccccc@{}}
			\toprule[1.2pt]
			& \multicolumn{3}{c}{Hough}    & \multicolumn{2}{c}{Sine}     & \multicolumn{2}{c}{Sine}     \\
			\multicolumn{1}{l}{} & \multicolumn{3}{c}{(SNR=10)} & \multicolumn{2}{c}{(SNR=10, $\alpha=0.3$)} & \multicolumn{2}{c}{(SNR=10, $\alpha=0.4$)} \\ \bottomrule
			& OL         & VC        & Time($s$)        & OL              & VC             & OL              & VC             \\ \bottomrule
			3rd-order  & 0.6        & 0.67      & 2        & 0.65            & 0.19           & 0.54            & 0.25           \\
			4th-order  & 0.71       & 0.84      & 2.4      & 0.62            & 0.48           & 0.56            & 0.33           \\
			5th-order  & 0.79       & 0.93      & 5.8      & 0.64            & 0.57           & 0.57            & 0.39           \\
			6th-order  & 0.85       & 0.96      & 9.2      & 0.66            & 0.67           & 0.62            & 0.82           \\
			SD\_Stream & 0.65       & 0.80       &          & 0.45            & 0.16           & 0.39            & 0.04           \\
			iFOD2      & 0.82       & 0.95      &          & 0.62            & 0.55           & 0.51            & 0.05           \\ \toprule[1.2pt]
		\end{tabular}
	}
\end{table}
\begin{figure}[]
	\centering
	\includegraphics[width=0.5\textwidth]{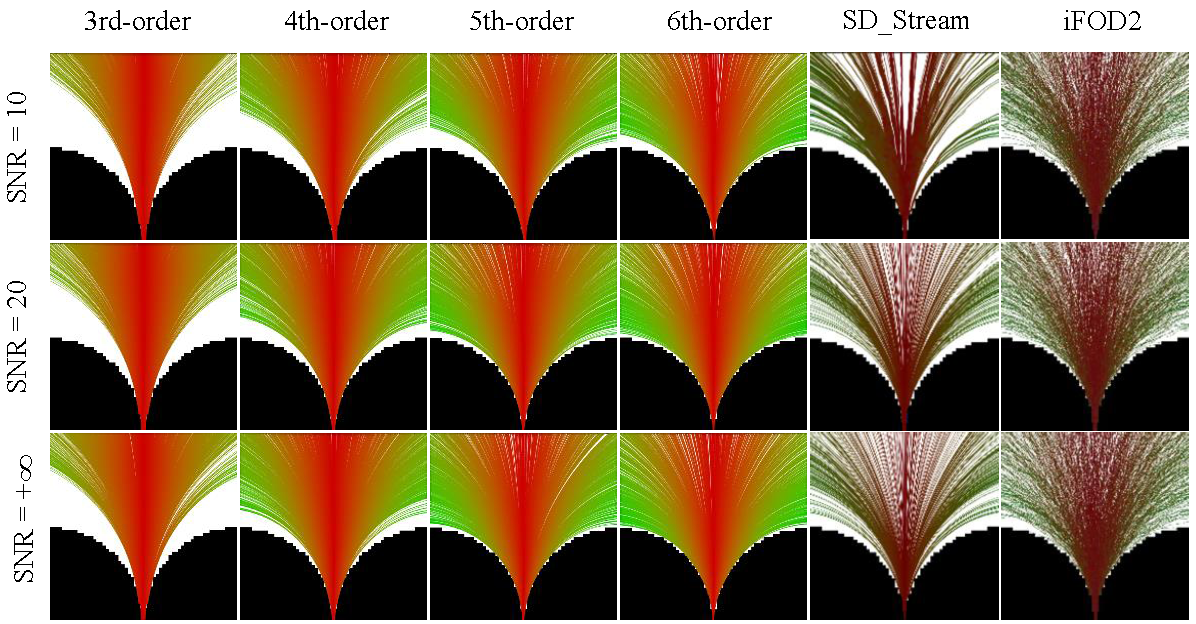}
	\caption{Comparison of tractography results of BTD with iFOD2 and SD\_Stream at different SNRs using the Hough data.}
	\label{fig:2}
\end{figure}
\begin{figure}[]
	\centering
	\includegraphics[width=0.48\textwidth]{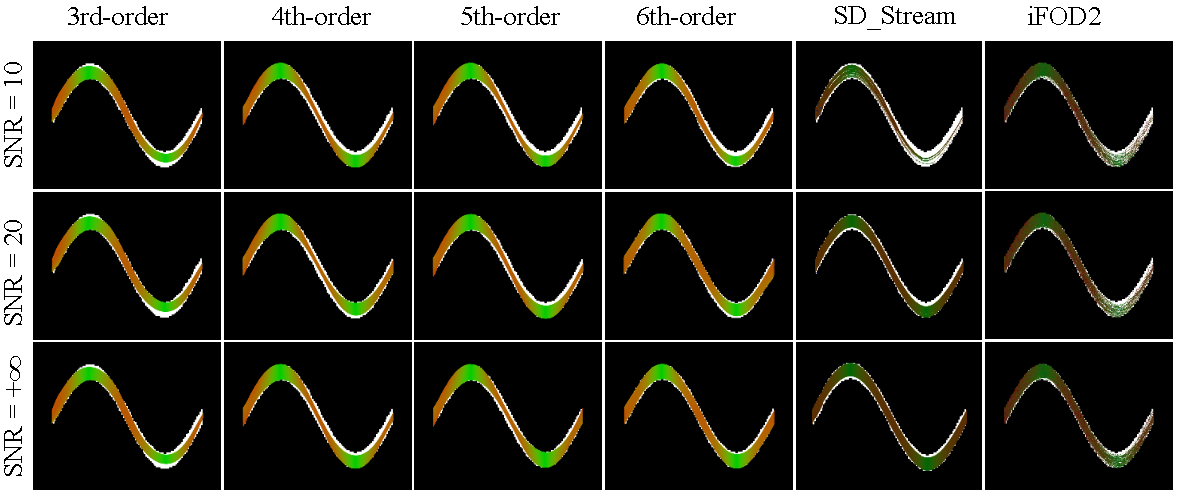}
	\caption{Comparison of tractography results of BTD with iFOD2 and SD\_Stream at different SNRs using the Sine data ($\alpha=0.3$).}
	\label{fig:3}
\end{figure}
\begin{figure}[h]
	\centering
	\includegraphics[width=0.48\textwidth]{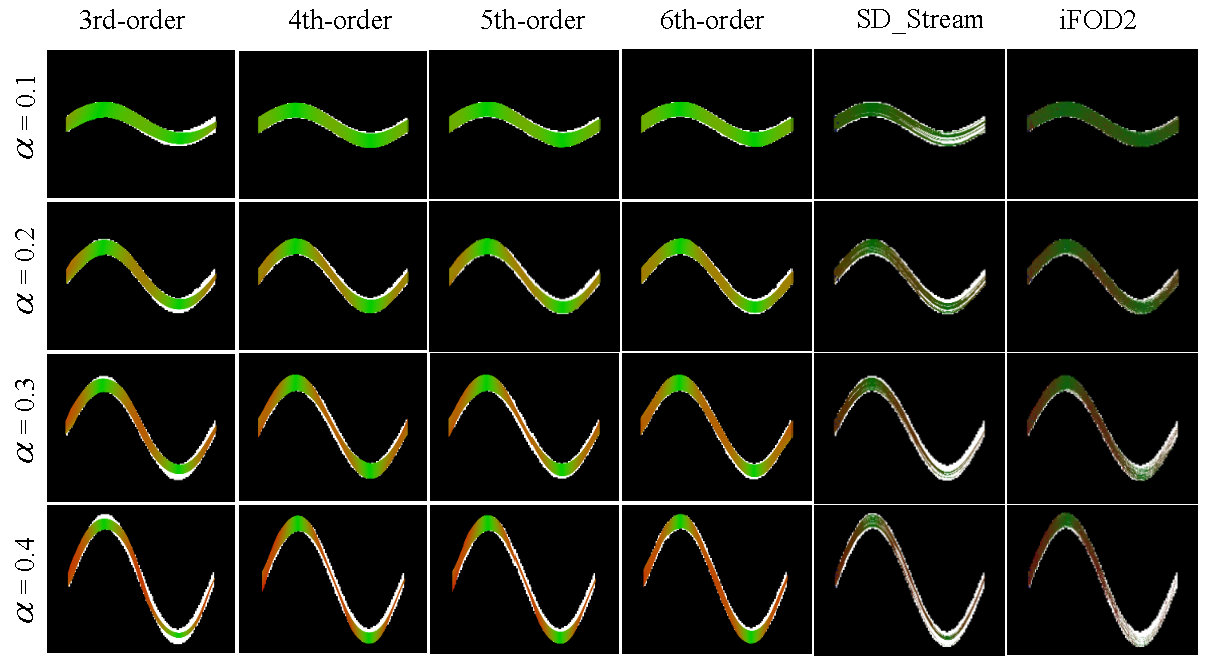}
	\caption{Comparison of tractography results of BTD with iFOD2 and SD\_Stream at different $\alpha$ using Sine data (SNR=10).}
	\label{fig:4}
\end{figure}
\begin{figure}[h]
	\centering
	\includegraphics[width=0.47\textwidth]{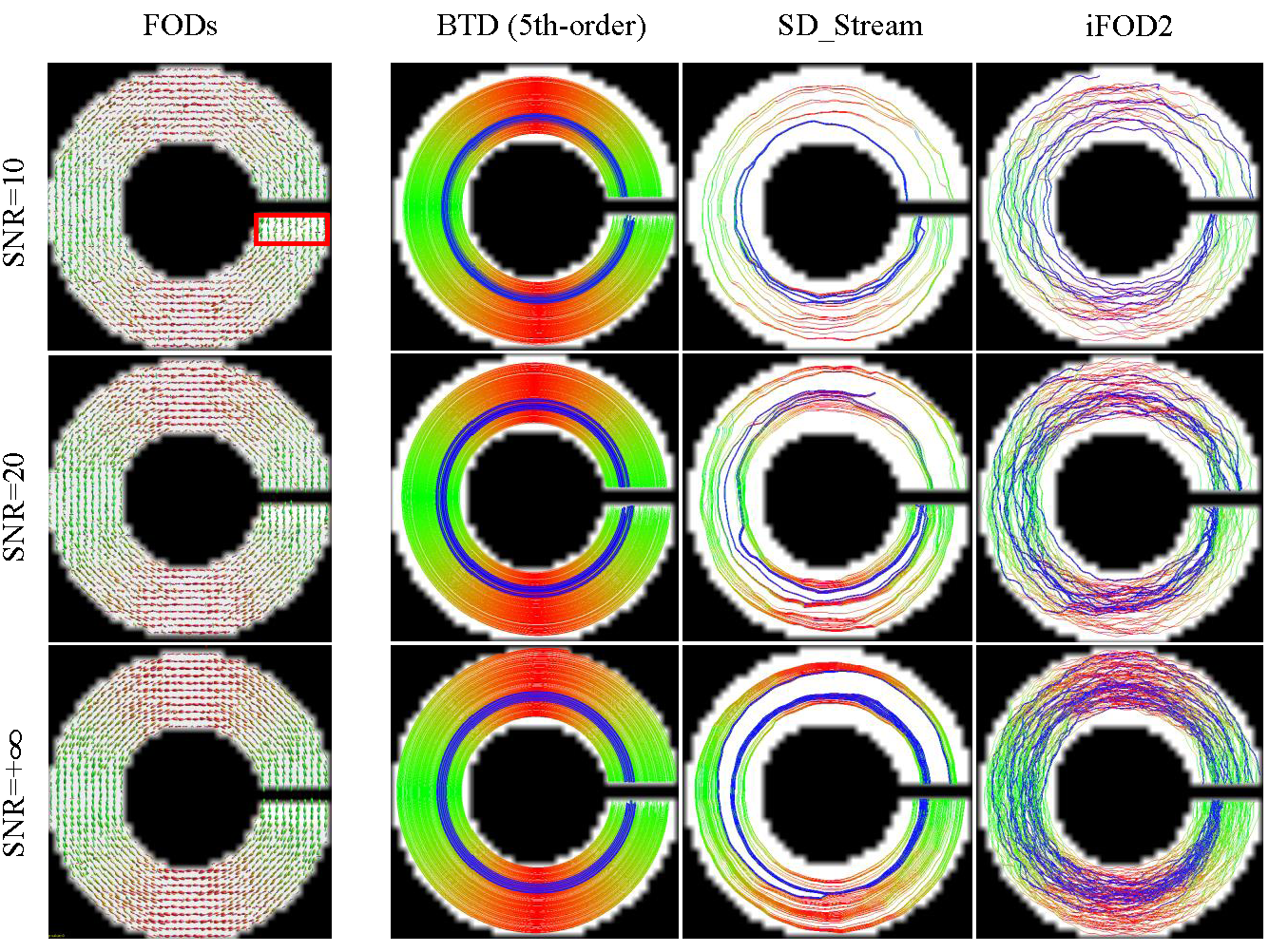}
	\caption{Comparison of tractography results of the BTD (5th-order) with iFOD2 and SD\_Stream at SNRs of 10, 20, and $+ \infty$ using Circle data. The blue color fibers (from the same starting area) are used to visualize the tractography deviation with different methods.}
	\label{fig:5}
\end{figure}
\begin{figure}[h]
	\centering
	\includegraphics[width=0.48\textwidth]{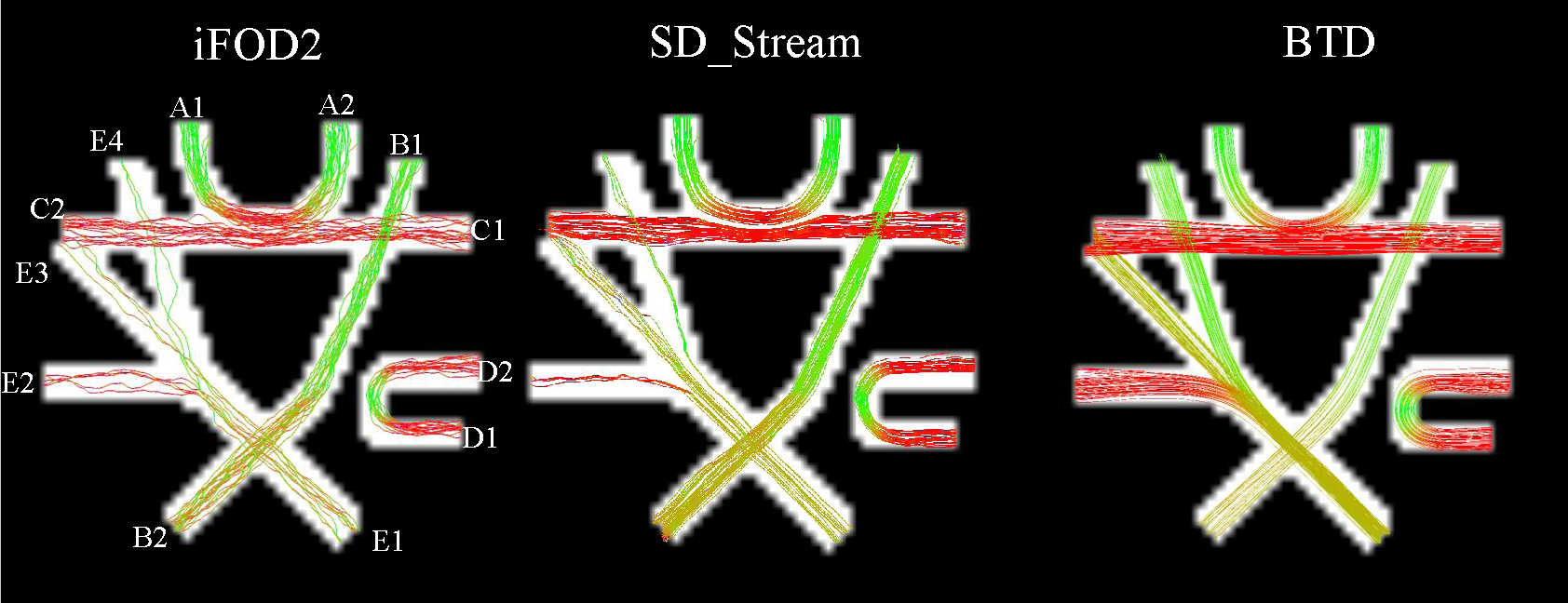}
	\caption{Comparison of tractography results of the BTD (5th-order) with iFOD2 and SD\_Stream using FiberCup data.}
	\label{fig:6}
\end{figure}

Tractography parameters are as follows: For Hough data, Sine data, Circle data, and FiberCup data: i) iFOD2: \textit{maximum angle} = $45^{0}$, \textit{step size} = 0.5, \textit{cutoff} = 0.1; ii) SD\_Stream: \textit{maximum angle} = $60^{0}$, \textit{step size} = 0.5, \textit{cutoff} = 0.1. The seed region for the Hough data is the first six rows of voxels (bottom of the data), the Sine data is the first two columns of voxels (left side of the data), and the Circle data is the four rows of voxels in the red box (Fig.~\ref{fig:5}). The seed masks are the first four columns or rows in regions A1, B1, C1, D1 and E1 for FiberCup data. The number of seeds is set to 2000 for the Sine and Hough data and 720 for the Circle data, and there are 4 seeds in each voxel for the FiberCup data. The minimum length of any track is set to 5$mm$ (5 voxels) for the Hough and Sine data, 15$mm$ (5 voxels) for the FiberCup data, and ${\rm{2}}\pi  \cdot {{\rm{r}}_1}$ mm for the Circle data.

\begin{table}[]
	\centering
	\caption{Comparison of Tractometer metrics of BTD using Circle data with SD\_Stream, and IFOD2.}
	\label{tab:2}
	\resizebox{0.45\textwidth}{!}{%
		\begin{tabular}{ccccc}
			\toprule[1.2pt]
			& SNR & 5th-order & SD\_Stream & iFOD2 \\ \bottomrule
			\multirow{3}{*}{VC}                & 10  & 0.47      & 0.21       & 0.30   \\
			& 20  & 0.60       & 0.28       & 0.40   \\
			& $+ \infty$   & 0.61      & 0.39       & 0.58  \\ \bottomrule
			\multirow{3}{*}{OL}                & 10  & 0.81      & 0.58       & 0.75  \\
			& 20  & 0.83      & 0.64       & 0.79  \\
			& $+ \infty$   & 0.90       & 0.62       & 0.85  \\ \bottomrule
			\multirow{3}{*}{Deviation (voxel)} & 10  & 0.48      & 0.88       & 1.61  \\
			& 20  & 0.42      & 1.31       & 1.71  \\
			& $+ \infty$   & 0.23      & 0.91       & 1.48 \\ \toprule[1.2pt]
		\end{tabular}
	}
\end{table}
\begin{table}[b]
	\centering
	\caption{Comparison of Tractometer metrics of BTD using FiberCup data with SD\_Stream, and IFOD2.}
	\label{tab:3}
	\resizebox{0.35\textwidth}{!}{%
		\begin{tabular}{ccccc}
			\toprule[1.2pt] 
			& Bundle       & BTD  & iFOD2 & SD\_Stream \\ \bottomrule
			\multirow{7}{*}{VC} & A1 A2        & 0.73 & 0.69  & 0.30       \\
			& B1 B2        & 0.54 & 0.48  & 0.32       \\
			& C1 C2        & 0.74 & 0.44  & 0.50       \\
			& D1 D2        & 0.73 & 0.64  & 0.33       \\
			& E1 E2        & 0.78 & 0.08  & 0.06       \\
			& E1 E3        & 0.46 & 0.10  & 0.15       \\
			& E1 E4        & 0.85 & 0.53  & 0.23       \\ \bottomrule
			\multirow{7}{*}{OL} & A1 A2        & 0.50 & 0.47  & 0.43       \\
			& B1 B2        & 0.48 & 0.45  & 0.43       \\
			& C1 C2        & 0.62 & 0.43  & 0.41       \\
			& D1 D2        & 0.46 & 0.32  & 0.39       \\
			& E1 E2        & 0.42 & 0.14  & 0.13       \\
			& E1 E3        & 0.35 & 0.19  & 0.30       \\
			& E1 E4        & 0.45 & 0.35  & 0.17       \\ \bottomrule
			\multirow{5}{*}{OR} & A1 A2        & 0.05 & 0.12  & 0.06       \\
			& B1 B2        & 0    & 0.02  & 0          \\
			& C1 C2        & 0.11 & 0.16  & 0.13       \\
			& D1 D2        & 0    & 0     & 0          \\
			& E1 E2, E3 E4 & 0    & 0.06  & 0.008   \\ \toprule[1.2pt] 
		\end{tabular}
	}
\end{table}

To evaluate the tractograms of BTD on different orders, we test BTD from the 3rd- to 6th-order on Hough data and Sine ($\alpha  = 0.3$) data with SNRs of 10, 20, and $+ \infty$, which are shown in Fig.~\ref{fig:2} and Fig.~\ref{fig:3}. We further test the algorithms on Sine data with $\alpha$ from 0.1, 0.2, 0.3, and 0.4 (Fig.~\ref{fig:4}) to adjust the amplitude. The quantitative results of Hough data with SNR=10, Sine data with $\alpha  = 0.3$ and SNR=10, and Sine data with $\alpha  = 0.4$ and SNR=10 are shown in Table.~\ref{tab:1}. From Fig.~\ref{fig:2}, Fig.~\ref{fig:3} and Table.~\ref{tab:1} show the fitting ability increase, and the BTD with 5th-order and 6th-order yield better results than the 3th-order and 4th-order BTD. Compared to the 5th-order BTD, the 6th-order BTD shows approximate fitting ability but the complexity will increase significantly because the coefficients of the BTD from the 5th-order to the 6th-order will increase by 84 terms. Therefore, the 5th-order BTD is used to compare the tractography results in the following experiments. Notion, we can select the lower order of BTD when we track the simple bundles, which can reduce running time. However, for the complex bundle, like the corpus callosum, we suggested the higher order of BTD as the bundle is complex. Moreover,  for most of the bundle in vivo, the fitting ability of 5th-order is sufficient and the run time is suitable. Therefore, we recommended the 5th-order of BTD for some complex bundles.
\begin{table}[t]
	\centering
	\caption{Comparison of Tractometer metrics of BTD using ISMRM 2015 tractography challenge data with UKF, SD\_Stream, and IFOD2.}
	\label{tab:4}
	\resizebox{0.43\textwidth}{!}{%
		\begin{tabular}{ccccccc}
			\toprule[1.2pt] 
			& \multicolumn{3}{c}{CST-L} & \multicolumn{3}{c}{CST-R} \\ \bottomrule
			& VC      & OL     & OR     & VC      & OL     & OR     \\ \bottomrule
			BTD        & 0.79    & 0.80   & 0.09   & 0.72    & 0.85   & 0.10   \\
			UKF        & 0.26    & 0.77   & 0.23   & 0.19    & 0.84   & 0.21   \\
			iFOD2      & 0.23    & 0.74   & 0.18   & 0.34    & 0.72   & 0.19   \\
			SD\_Stream & 0.10    & 0.57   & 0.11   & 0.16    & 0.53   & 0.13  \\ \toprule[1.2pt] 
		\end{tabular}
	}
\end{table}
\begin{figure}[]
	\centering
	\includegraphics[width=0.43\textwidth]{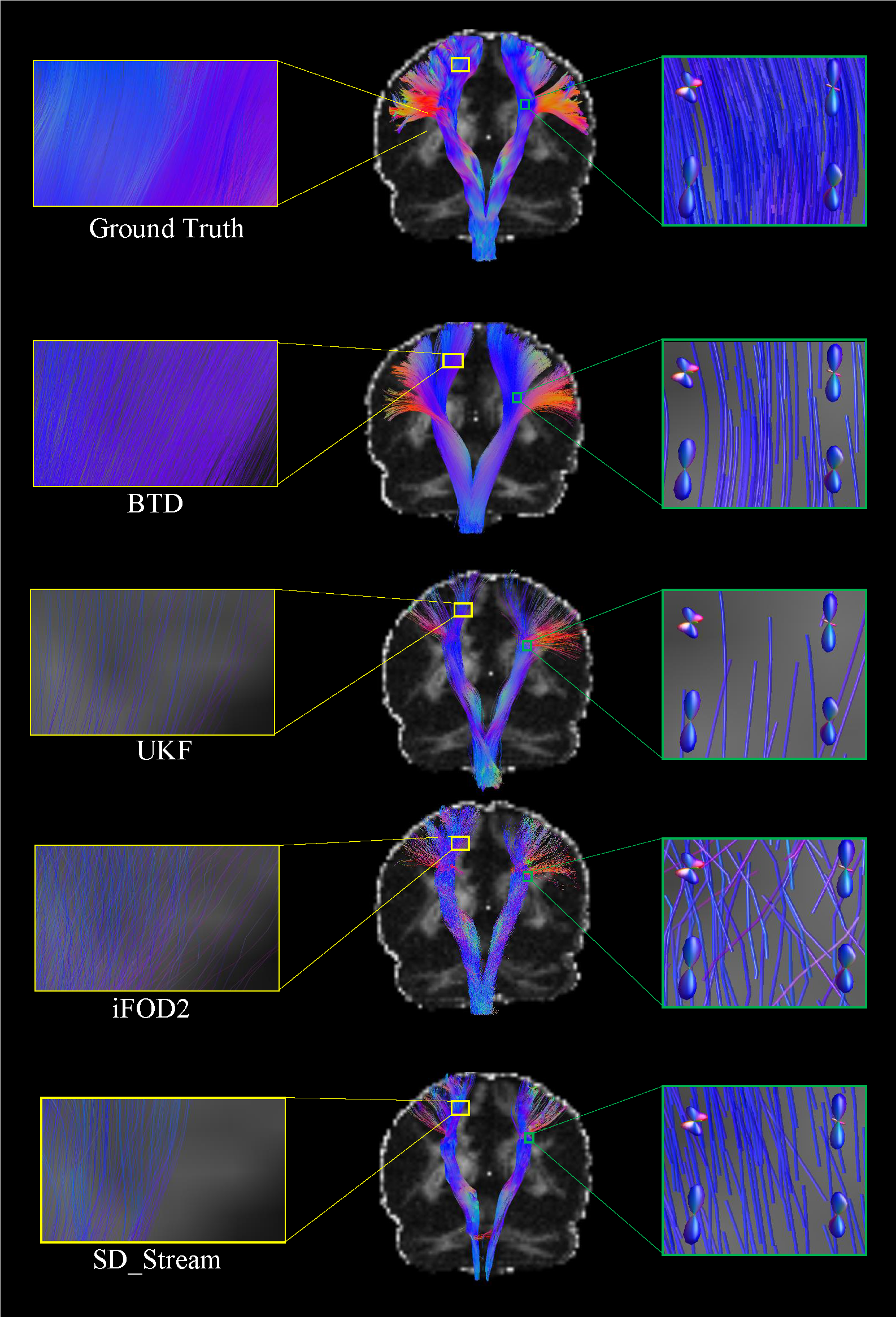}
	\caption{Comparison of tractography results of BTD with UKF, iFOD2, and SD\_Stream on CST using ISMRM 2015 Tractography Challenge data.}
	\label{fig:7}
\end{figure}
\begin{figure}[]
	\centering
	\includegraphics[width=0.45\textwidth]{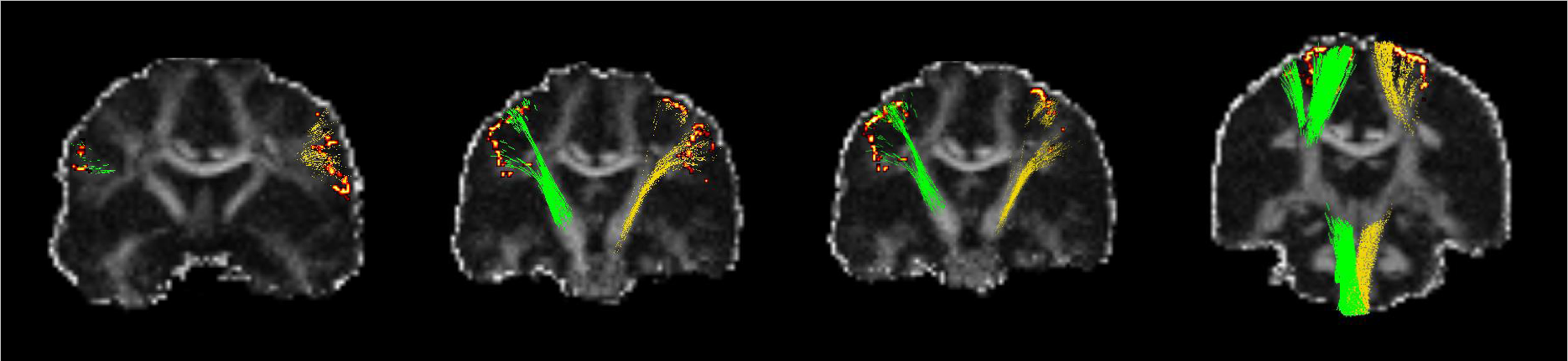}
	\caption{Slices and fibers of BTD on ISMRM 2015 Tractography Challenge data.}
	\label{fig:8}
\end{figure}

To further verify the proposed algorithm, we assess the tractograms using Circle data with SNRs of 10, 20, and $+ \infty$, in which error accumulation is more obvious. The results are shown in Fig.~\ref{fig:5} and Table.~\ref{tab:2}. The blue curves represent the fibers in the same starting area but with different trajectories. The ending points of the SD\_Stream and iFOD2 deviate largely compared to their starting points. 

To illustrate the model performance in one phantom image covering various streamline scenarios, we test BTD on FiberCup data. This data included crossing, fanning etc., which was widely used for comparison of tracking algorithms. The BTD has better performance for the bundles with crossing and twisting regions in Fig.~\ref{fig:6}.  Furthermore, the BTD shows better VC, OL and OR in Table.~\ref{tab:3}.

We compare the tractograms among 5th-order BTD, SD\_Stream, and iFOD2. From Fig.~\ref{fig:2}, the tractograms of BTD are evenly distributed in the mask and have larger VC and OL than SD\_Stream and iFOD2 with different SNRs. In addition, the tractograms of SD\_Stream and iFOD2 show the small angle of divergence, and fewer fibers reach the large fanning regions on Hough data. As an important factor affecting tractography, error accumulation leads to premature termination of the fibers, which is more obvious in long-range and large twisting connections, such as the bundle on Sine data (Fig.~\ref{fig:3}). The BTD obtains larger spatial coverage as well as better VC and OL at different SNRs. To further compare the tractograms on more complex data, we adjust the  from 0.1, 0.2, 0.3, and 0.4 for Sine data (Fig.~\ref{fig:4}). The BTD shows more stable tractograms and higher VC and OL, while SD\_Stream and iFOD2 exhibit an increase in the number of prematurely terminated fibers with decreasing amplitude.  In Fig.~\ref{fig:5}, the BTD shows less deviation with increasing noise and most of fibers can return their starting points. In Fig.~\ref{fig:6} and Table.~\ref{tab:3}, the BTD shows better performance compared with SD\_Stream and iFOD2, specifically for the bundles with crossing and fanning (E1-E2, E1-E4). The BTD shows large VC and OL and lower OR compared with SD\_Stream and iFOD2. Additionally, the computational time from 3th- to 6th-order BTD may be need approximately 2.0s, 2.4s, 5.8s and 9.2s runtime using Hough data (repeat 100 times; in fourth column in Table.~\ref{tab:1}) with Inter i-9900k processor and Matlab2019 platform.  

From the above results, the BTD has more valid fibers, larger spatial coverage, and lower error accumulation as fibers spread forward than SD\_Stream and iFOD2. The BTD seems to capture the better fanning, long-range and twisting, and large bending bundle tracking results.  

\subsection{ISMRM 2015 Tractography Challenge data}
In this section, we evaluate the performance of the BTD on the ISMRM 2015 Challenge data, which simulates the shape and complexity of 25 well-known in vivo fiber bundles. The dataset has 32 gradient directions, a b-value of 1000 $s/m{m^2}$, and 2$mm$ isotropic voxels. The dataset is denoised and corrected for distortions using MRtrix3 ($dwidenoise$ and $dwipreproc$). The tractography parameters are as follows: i) iFOD2: \textit{maximum angle} = $30^{0}$, \textit{step size} = 0.2, \textit{cutoff} = 0.1, \textit{minimum length} = 50; ii) SD\_Stream: \textit{maximum angle} = $60^{0}$, \textit{step size} = 0.3, \textit{cutoff} = 0.1, \textit{minimum length} = 75; iii) UKF: \textit{seedingFA} = 0.06, \textit{stoppingFA} = 0.05, \textit{stoppingThreshold} = 0.06, \textit{Qm} = 0.001, and \textit{Ql} = 50. The start and end point of our proposed BTD use the officially provided ground truth. FODs are performed by standard constrained spherical deconvolution (CSD) in MRtrix for iFOD2 and SD\_Stream, and the UKF uses a two-tensor model. 

We selected the corticospinal tract (CST) as an example to test the algorithms. The CST has the features of large fanning and long range. The anatomy of CST is well known from the brainstem to the precentral gyrus~\cite{52}. The bundle masks are the voxels that ground truth fibers pass through after dilatation. In Fig.~\ref{fig:7}, we exhibit the details of the CST near area 4t (yellow boxes) and another regions (green boxes)(Brainetome regions~\cite{54}). Tractometer metrics with VC, OR, and OL for left and right CST are presented in Table.~\ref{tab:4}.

\begin{figure*}[]
	\centering
	\includegraphics[width=0.99\textwidth]{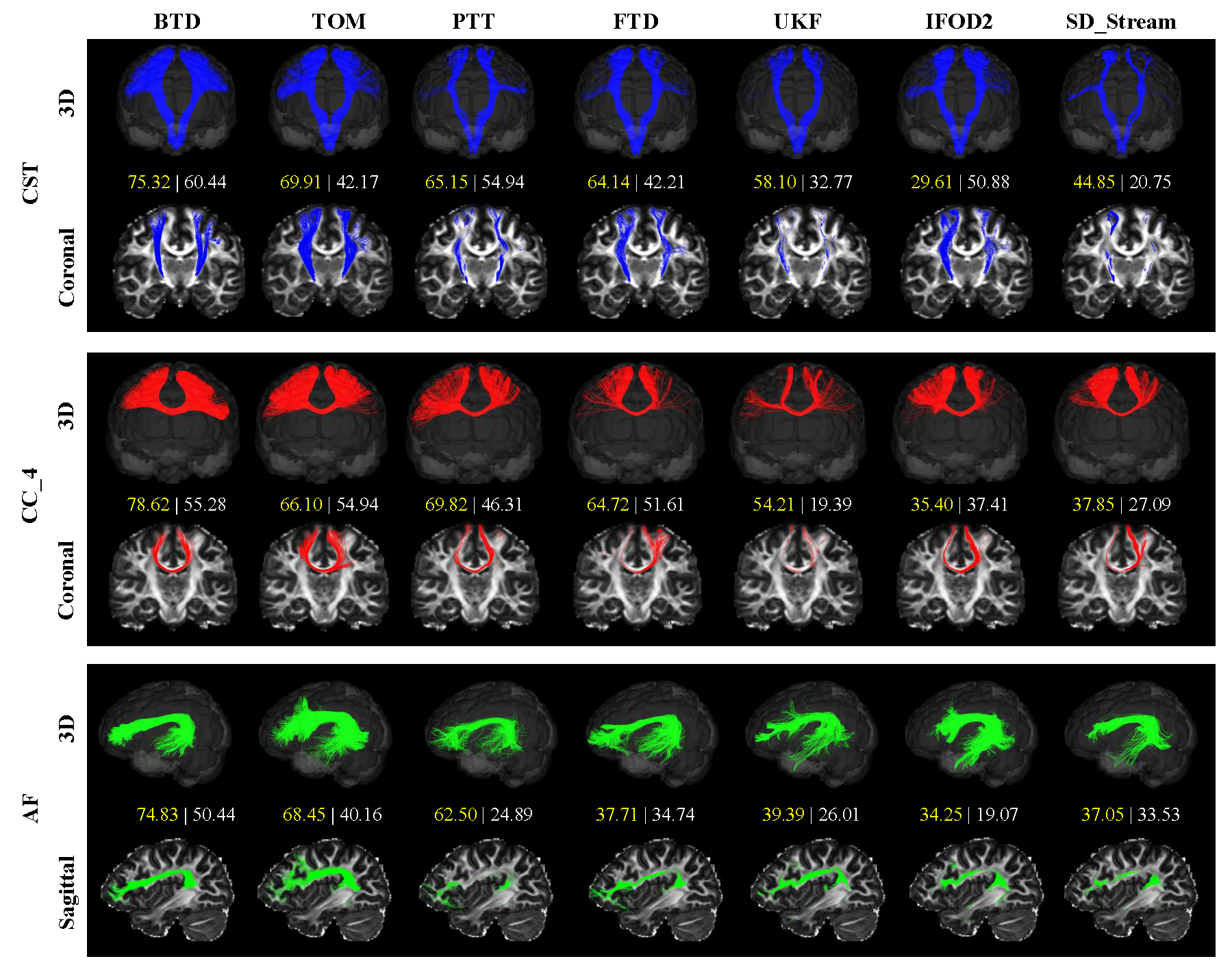}
	\caption{Qualitative comparison of tractography results reconstruction of CST, CC\_4, and AF on HCP \#100307 subject. Top: the 3D display of tracked fiber bundles. Bottom: the slice display of tracked fiber bundles on the FA images. Middle: the quantitative comparison of different tractography methods with tractometer metrics (VC(\%) $|$ OL(\%)).}
	\label{fig:12}
\end{figure*}
From the right column in Fig.~\ref{fig:7}, the BTD preserves better spatial fluency and is closer to the ground truth. The BTD tractography more fibers ending in precentral gyrus than iFOD2, SD\_Stream, and UKF methods.  While the UKF shows some twisted fibers and is unevenly distributed. The iFOD2 and SD\_Stream show sparse and interrupted fibers. The BTD can track the large fanning fibers that ending nearby 4tl and 4hf in precentral gyrus. The iFOD2 and SD\_Stream show fewer or no fibers in these regions. We can see that the VC and OL of iFOD2 and SD\_Stream in Table.~\ref{tab:4} are lower than BTD and UKF. In addition, the BTD has a lower OR compared with other three algorithms. The BTD seems to capture the complexity in regions where we expect fiber geometry (details are shown in Fig.~\ref{fig:8}). This is because the BTD reconstructs a bundle in a ‘cluster to cluster’ manner to reduce the ambiguous spatial correspondences between diffusion directions and fiber geometry. Therefore, the BTD preserves better spatial fluency and can better track the complex fibers than current peak-based tractography.
\subsection{HCP data}\label{sec:HCP}
For visual and quantitative comparisons on data from real subjects, we used the HCP dataset subjects~\cite{55}. These are acquired using 288 gradient directions, consisting of 18 scans at b = 0 $s/m{m^2}$ and three b-values (1000 $s/m{m^2}$, 2000 $s/m{m^2}$, 3000 $s/m{m^2}$) using 90 gradients, and the voxel size is 1.25$mm$ × 1.25$mm$ × 1.25$mm$. We used the preprocessed dMRI images shared by HCP. FODs were  estimated using constrained spherical deconvolution, which are included in the software package MRtrix~\cite{10}. 

We used HCP \#100307 subject to visually compare fiber tractograms from the proposed BTD algorithm and results from the tract orientation mapping (TOM) \cite{58}, parallel transport tractography (PTT) \cite{30}, fiber trajectory distribution (FTD)~\cite{31}, unscented Kalman filter (UKF) algorithm~\cite{47}, integration over fiber orientation distributions (iFOD2)~\cite{57}, and deterministic FOD-based tracking (SD\_Stream)~\cite{56}. In this paper, we use VC and OL to validate the proposed method, so all tractography methods were selected 3000 streamlines which were seeded from all voxels within the each tract start regions, and tract masks were used to filter the tractograms. The tractography specific parameters are as follows: i) FTD, iFOD2, and SD\_Stream: \textit{maximum angle} = $60^{0}$, \textit{step size} = 0.3, \textit{cutoff} = 0.1, \textit{minimum length} = 75; ii) UKF: \textit{seedingFA} = 0.06, \textit{stoppingFA} = 0.05, \textit{stoppingThreshold} = 0.06, \textit{Qm} = 0.001, and \textit{Ql} = 50. iii) PTT: the default parameters in \cite{30}, the difference is that the seeding regions are the tract start regions. iv) TOM: the default parameters in \cite{58}, the difference is that the seeding regions are the tract start regions, \textit{max\_nr\_fibers} = 3000, no streamline filtering by end mask. For the HCP data, the start and end point of our proposed BTD use the start and end region segmented by the TractSeg~\cite{wasserthal2018tractseg} method. To ensure the fairness of the experiment, all competing methods use the same start point, end point and the bundle mask.

In the absence of ground truth fibers for in vivo data, the relatively familiar corpus callosum (CC\_4), CST, and arcuate fasciculus (AF) were selected for qualitative and quantitative evaluations. The AF is a neuronal pathway that connects Wernicke’s area and Broca’s area~\cite{51}. The CC\_4 is minor forceps of the corpus callosum that connects bilateral frontal lobe. These three tracts have the characteristics of long-range, twisting, and fanning characteristics, making them suitable to assess the algorithms. 
\begin{figure}[t]
	\centering
	\includegraphics[width=0.48\textwidth]{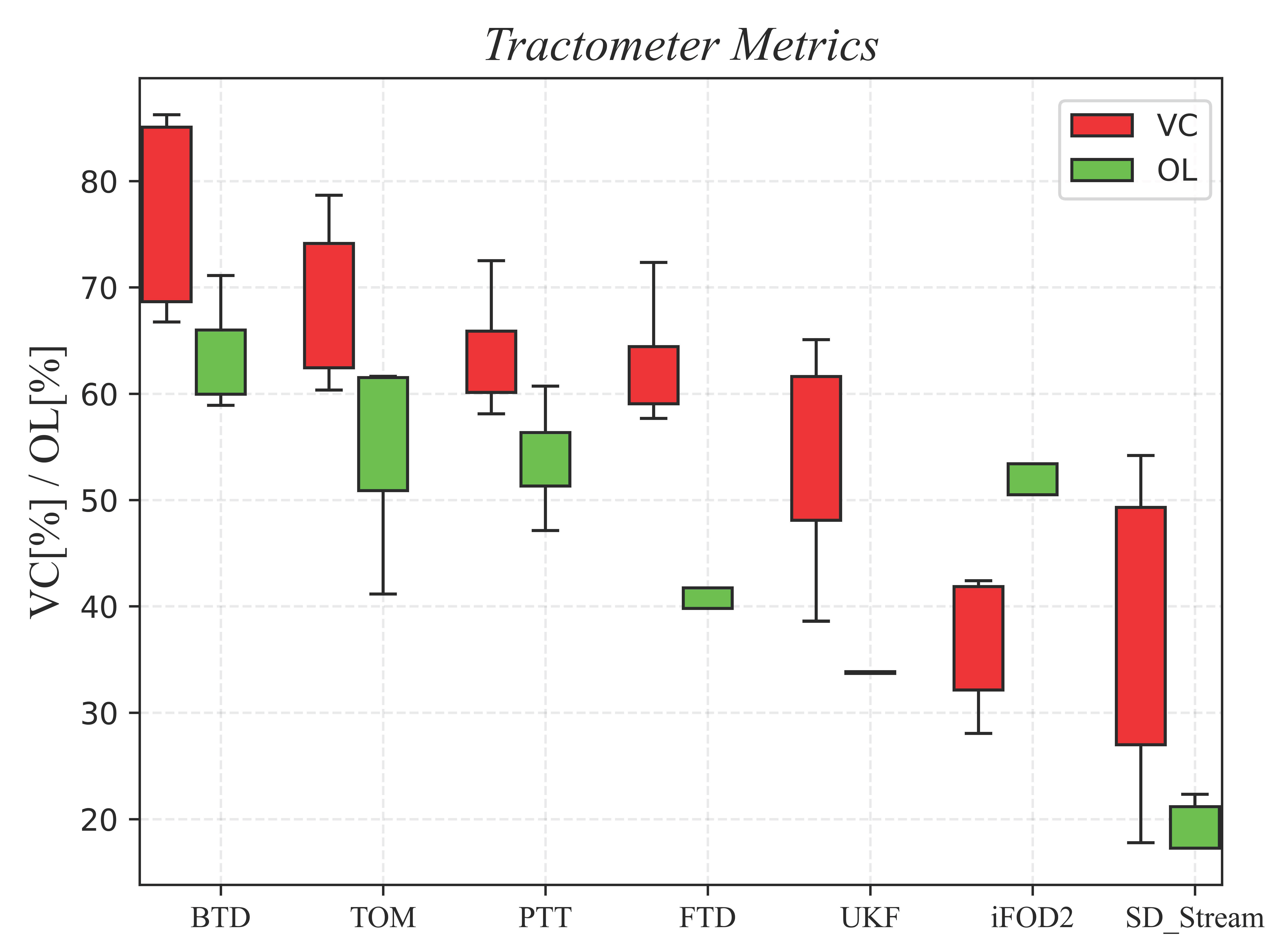}
	\caption{Box-plots of tractometer metrics comparing reconstruction of CST using the proposed BTD with other tractography methods.}
	\label{fig:13}
\end{figure}

The main lost fibers in the TOM, PTT, FTD, UKF, iFOD2, and SD\_Stream are mainly distributed in fanning and other complex geometric structures, such as the fibers ending in area 4tl and 4hf of precentral gyrus (CST), frontal lobe on CC\_4 and temporal lobes on AF. The results are consistent with VC and OL in middle of Fig.~\ref{fig:12}, in which the BTD obtains highest VC and OL. We also exhibited the fibers of AF and CC\_4 on anatomical slices in top of Fig.~\ref{fig:12} (the slices and fibers of CC\_4 can be seen in bottom of Fig.~\ref{fig:12}). In addition, the BTD of 7th-order or higher may overfit in the termination region. 

In addition, we also give tractometer metrics results of the proposed BTD on CST
using five HCP data (subject ID: \#100307, \#112112, \#112920, \#113821, \#118831) in Fig.~\ref{fig:13}. The results show that the proposed method is higher than the other compared methods in VC and OL tractometer metrics. We can also see that BTD has a significant improvement compared to FTD, which illustrates the advantage of BTD for complex bundle reconstruction by establishing higher-order streamline differential equations at the  global level.

\section{Discussion}\label{sec:Discussion}
In this paper, we define a bundle-specific tractogram distribution (BTD) function based on any higher-order streamline differential equation in the measured diffusion vectorial field. The results across the six different datasets show that the proposed BTD is capable of reconstructing complex fiber bundles with long distances, large twists, and fan-shaped bundles, and shows better spatial consistency with the fiber geometry. The order is an important parameter in our proposed method, so we give the effect of different orders on the proposed BTD and from the experimental results we can see that the 5th- or 6th-order BTD is the better parameter. Additionally, we also tested the computational speed required by the proposed algorithm and the results show that our method is reasonable time. Last but not least, our approach can accomplish fiber reconstruction without the need for expert ROIs placement, which also lowers inter-operator bias.

In our previous works~\cite{31}, we designed the fiber trajectory distribution (FTD) function defined on the neighborhood voxels by using a ternary quadratic polynomial-based streamline differential equation, which can reveal continuous asymmetric fiber trajectory. However, the FTD is still a local reconstruction of the fiber trajectory at the neighborhood voxel level. Different from the FTD function, the proposed BTD formulates the tractography problem in the Riemannian manifold and integrates higher-order streamline differential equations to derive bundle-specific tractography. The BTD does not just is to fit a divergence-free vector field, but is simplified as the estimation of BTD coefficients by minimizing the energy optimization model, and is used to characterize the relations between BTD and diffusion tensor vector under the prior guidance by introducing the tractogram bundle information to provide anatomic priors. There is a reason why the proposed method provides such a good performance. The proposed BTD is to convert the fiber bundle connectivity relations modeled in terms of a voxelwise or local into the flow of the overall fluid on the Riemannian manifold. In the BTD function, the tangent vector at each point of the fiber path equals to field vector of the diffusion tensor, which is consistent with the definition of fiber tracking~\cite{1,4}. The BTD function estimation algorithm takes pre-computed peak directions, voxel centroids, seed points, and tractography steps as input, and results in a bundle-specific tractogram.

In this paper, we define a new BTD function to describe the tract-wise tractogram. In order to ensure the fairness of the experiment, the bundle mask used in all our experiments was kept the same as the TOM method~\cite{wasserthal2018tractseg}~\cite{58}. For a given bundle mask, the proposed BTD function uses it initially for localization, then aims for global optimization within this localized region. Since all the comparison methods have slightly different approaches and requirements, for a fair comparison, we impose some restrictions on the other methods (TOM \cite{58}, PTT \cite{30}, FTD~\cite{31}, UKF~\cite{47}, iFOD2~\cite{57}, and SD\_Stream~\cite{56}) when validating on vivo data from the Human Connectome Project (HCP) data, with the detailed parameters in Section~\ref{sec:HCP}. In this paper, we use VC and OL to validate the proposed method, so we set all comparison experiments to set the same seeds in the same start region, using the mask as a regional restriction for fiber tracking to compare the results. There is experimental evidence~\cite{1}~\cite{cote2013tractometer} that complete seeding on the mask and the ROIs have an impact on the performance of fiber tracking. Therefore, there are some differences between the TOM we reproduced and the original results.

Tractography algorithms that offer better tools for examining brain topographic patterns are beneficial for both clinical and fundamental science applications. By integrating higher-order streamline differential equations, the proposed BTD poses the tractography issue in Riemannian manifolds and results in the derivation of bundle-specific tractography. We believe that the proposed BTD opens up new possibilities for structural brain mapping. We think that as neural network techniques progress, BTD can be used to provide a more accurate trajectory flow for any traction approach, not just for direct fiber bundle trajectory tracking but also as a direction via the fiber bundle region as a priori.

The following are some possible limitations of this study as well as some potential future research to solve these problems. Firstly, errors in the FOD or peaks of a single voxel will not affect our proposed method, but it will be affected when there are numerous errors due to noise, artifacts, pathological conditions, or low-quality datasets. More work is needed to directly utilize the original diffusion MRI signals instead of peaks in the fiber orientation distribution to fit the flow field direction. Secondly, although an exact segmentation of fiber bundles is not necessary, the presence of multiple fiber bundles within a single mask will affect the proposed method. Thus, the future study will therefore focus on figuring out how to model the ideal transport on the flow field and how to address the issue of false positive fibers showing up during the tracking process.

\section{Conclusion}\label{sec:Conclusion}
In this work, a novel bundle-specific tractography approach BTD is proposed, which integrates higher-order streamline differential equation to derive brain connectome between two regions. We parameterize fiber bundles using the BTD coefficients that are estimated by minimizing the energy on the diffusion vectorial field by combining the priors. Experiments are performed on Hough, Sine, Circle data, FiberCup data, the ISMRM 2015 Tractography Challenge data, and in vivo data of HCP for qualitative and quantitative evaluation. The horizontal comparisons show that with increasing order of the BTD, the numbers of valid fibers and overlapped regions will gradually increase. The best results are obtained with the 5th- or 6th-order BTD, and an order higher than six may cause overfitting. The comparisons with state-of-the-art methods show that the BTD can reconstruct complex fiber bundles, such as long-range, large twisting, and fanning tracts, and show better spatial consistency with fiber geometry, which is potentially useful for robust tractography. 
\section*{Data and code availability}
The FiberCup data and the ISMRM 2015 Tractography Challenge data are available at ~\href{https://tractometer.org/}{https://tractometer.org/}. The Human Connectome Project dataset is available at ~\href{https://db.humanconnectome.org}{https://db.humanconnectome.org}. The code of our method is openly available at~\href{https://github.com/IPIS-XieLei/BTD-Tractography}{https://github.com/IPIS-XieLei/BTD-Tractography}.
\section*{Declarations of Interest}
The authors declare that they have no known competing financial interests or personal relationships that could have appeared to influence the work reported in this paper.
\section*{Acknowledgments}
This work was sponsored in part by the Zhejiang Province Science and Technology Innovation Leading Talent Program (No. 2021R52004), National Natural Science Foundation of China (Grant No.U22A2040, U23A20334, and 62303413) and Natural Science Foundation of Zhejiang Province (Grant No. LQ23F030017).

\bibliographystyle{IEEEtran}
\bibliography{IEEEabrv,myreference}

\end{document}